\documentclass[10pt,twocolumn,letterpaper]{article}

\usepackage{iccv}
\usepackage{times}
\usepackage{epsfig}
\usepackage{graphicx}
\usepackage{amsmath}
\usepackage{amssymb}

\usepackage[ruled]{algorithm2e}
\usepackage{booktabs}
\usepackage{multirow}

\usepackage[breaklinks=true,bookmarks=false]{hyperref}

\iccvfinalcopy % *** Uncomment this line for the final submission

\def\iccvPaperID{****} % *** Enter the ICCV Paper ID here

\begin{document}

%%%%%%%%% TITLE
\title{Progressive Cluster Purification for Transductive Few-shot Learning}

%% omit the following lines up until the closing ``}''.
%% Additional authors and addresses can be added with ``\and'',
%% just like the second author.
%% To save space, use either the email address or home page, not both
%\and

\author{Chenyang Si$^{1,2}$\thanks{Equal contribution.} \and Wentao Chen$^{1,3*}$ \and Wei Wang$^{1,2}$\thanks{Corresponding Author.} \and Liang Wang$^{1,2}$ \and Tieniu Tan$^{1,2,3}$
    \and
    $^1$Institute of Automation, Chinese Academy of Sciences (CASIA)\\
    $^2$University of Chinese Academy of Sciences (UCAS)\\
    $^3$University of Science and Technology of China (USTC)\\
    {\tt\small \{chenyang.si, wentao.chen\}@cripac.ia.ac.cn, \{wangwei, wangliang, tnt\}@nlpr.ia.ac.cn}
}

\maketitle
%\thispagestyle{empty}

%%%%%%%%% ABSTRACT
\begin{abstract}
    Few-shot learning aims to learn to generalize a classifier to novel classes with limited labeled data. Transductive inference that utilizes unlabeled test set to deal with low-data problem has been employed for few-shot learning in recent literature. Yet, these methods do not explicitly exploit the manifold structures of semantic clusters, which is inefficient for transductive inference. In this paper, we propose a novel Progressive Cluster Purification (PCP) method for transductive few-shot learning. The PCP can progressively purify the cluster by exploring the semantic interdependency in the individual cluster space. Specifically, the PCP consists of two-level operations: inter-class classification and intra-class transduction. The inter-class classification partitions all the test samples into several clusters by comparing the test samples with the prototypes. The intra-class transduction effectively explores trustworthy test samples for each cluster by modeling data relations within a cluster as well as among different clusters. Then, it refines the prototypes to better represent the real distribution of semantic clusters. The refined prototypes are used to remeasure all the test instances and purify each cluster. Furthermore, the inter-class classification and the intra-class transduction are extremely flexible to be repeated several times to progressively purify the clusters. Experimental results are provided on two datasets: miniImageNet dataset and tieredImageNet dataset. The comparison results demonstrate the effectiveness of our approach and show that our approach outperforms the state-of-the-art methods on both datasets.
\end{abstract}

%%%%%%%%% BODY TEXT
\section{Introduction}

\begin{figure}[t]
    \begin{center}
    \includegraphics[width=1.0\linewidth]{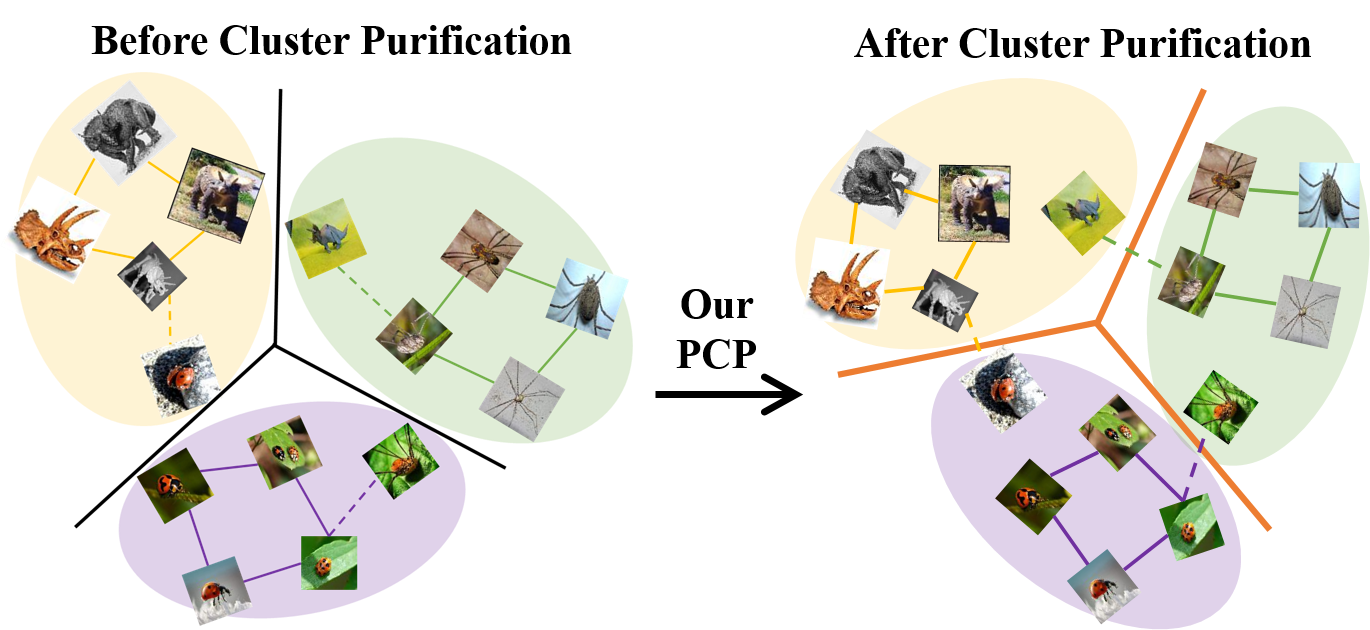}
    \end{center}
    \caption{For transductive few-shot learning, we first partition all query samples into several clusters. Due to the scarcity of labeled samples, there are a small number of samples with different semantics in each cluster. We propose a progressive cluster purification (PCP) method to progressively purify clusters. The PCP explicitly
    %models data relations among query samples to
    explore the manifold structure of each semantic cluster for transduction.}
\label{fig:cover}
\end{figure}

Deep learning~\cite{lecun2015deep} has enabled unprecedented achievements in various domains, such as computer vision~\cite{lin2014microsoft}, natural language processing~\cite{vaswani2017attention} and data mining~\cite{kipf2016semi}. These achievements are due to the improvement of algorithms and model architectures~\cite{krizhevsky2012imagenet,simonyan2014very,he2016deep,mikolov2013distributed}. However, remarkable performance often depends heavily on large amounts of labeled data to train the deep learning model, which is severely limited in practical application. In contrast, human visual system can recognize new classes from a few instances~\cite{lake2011one}. Therefore, it has attracted a lot of interests in extending the ability of deep learning methods for few-shot learning.~\cite{finn2017model,wang2018low,schwartz2018delta,vinyals2016matching}

\begin{figure*}[t]
\begin{center}
\includegraphics[width=0.9\linewidth, height=0.35\linewidth]{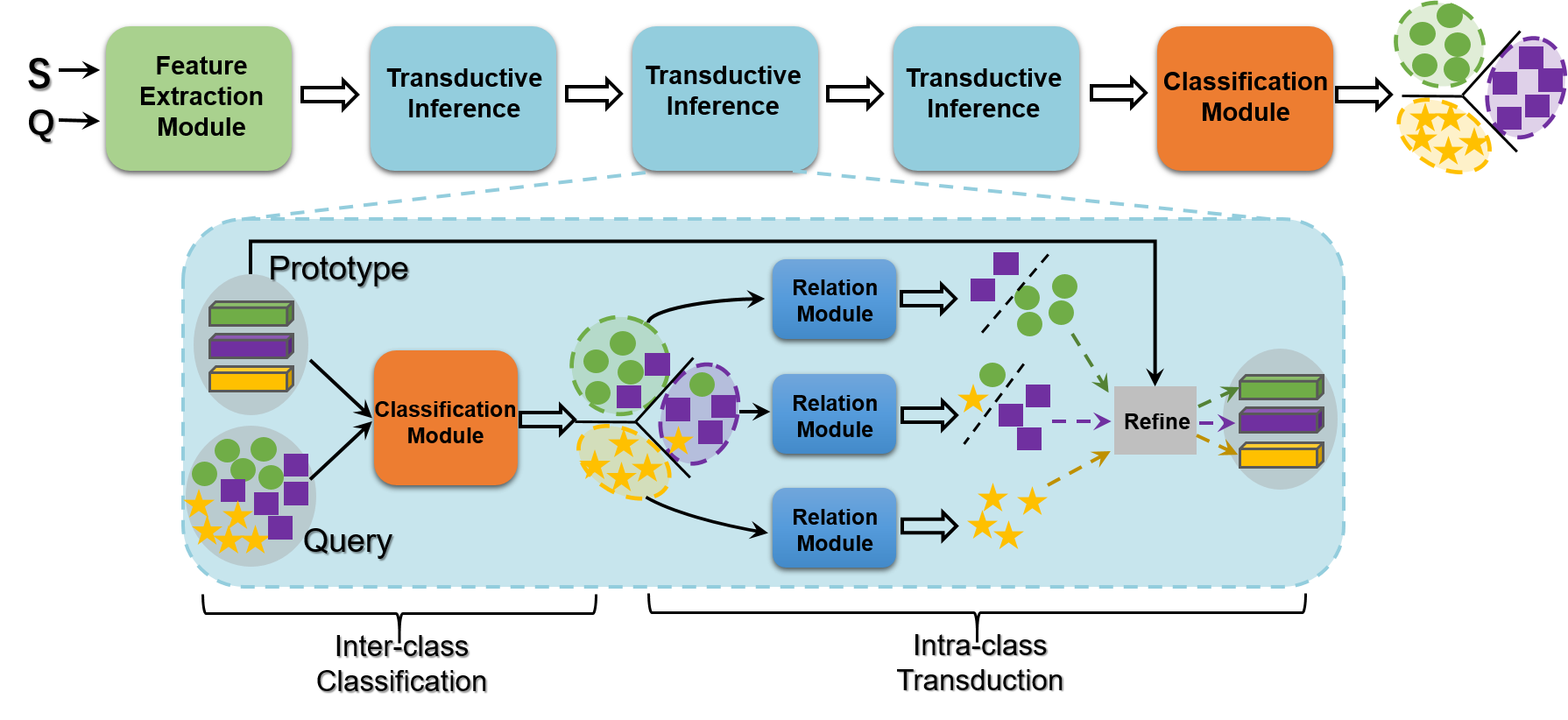}
\end{center}
   \caption{The architecture of the proposed Progressive Cluster Purification (PCP) method. We first extract visual features for input images and produce
   prototypes. Then, we partition all query samples into several clusters with a inter-class clustering model. For intra-class transduction, we explicitly model data relations between samples and select trustworthy samples to refine the prototype. The refined prototypes are used to remeasure all the query samples and purify each cluster. Furthermore, the inter-class clustering and the intra-class transduction are extremely flexible to be repeated for several times to progressively purify the clusters.}
\label{fig:Architecture}
\end{figure*}

Few-shot learning aims to learn to generalize a classifier to novel classes with limited labeled data. Although fine-tuning~\cite{jia2014caffe} has been widely used to transfer knowledge from large datasets, it is infeasible to obtain a generalized classifier by fine-tuning it with few labeled instances. Recent work tries to solve this problem from the perspective of meta-learning~\cite{lemke2015metalearning}, which extracts task-agnostic knowledge from a sequence of similar tasks and applies it to a new task for fast adaption only using few training data. There are several meta-learning methods for few-shot learning, such as learning sensitive and general initial parameters~\cite{finn2017model}, a learnable optimizer~\cite{ravi2016optimization} and a message passing network~\cite{mishra2017simple}. Another pipeline of work derives from metric learning~\cite{yang2006distance}, which directly measures the unseen samples based on the distance from labeled samples. For example, Vinyals \etal~\cite{vinyals2016matching} employ a simple cosine distance metric in embedding space to measure the similarity between two samples. The prototypical networks~\cite{snell2017prototypical} extract a prototype from few labeled samples to represent a new concept, and then consider it as a classifier to recognize new samples based on Euclidean distance. In addition, a straightforward idea to deal with data scarcity for few-shot learning is to synthesize new samples based on given data with generative adversarial networks (GANs)~\cite{goodfellow2014generative,zhang2018metagan} or auto-encoders~\cite{wang2018low,schwartz2018delta}.

Although these approaches are effective to generalize a classifier to novel classes, it is still challenging to learn a well-generalized semantic space with scarce data. Ren \etal~\cite{ren2018meta} propose novel extensions of the prototypical networks, which use external unlabeled examples to facilitate prototype production. However, a large number of unlabeled samples from different classes will interfere with the production of more trustworthy prototypes. Inspired by transductive inference~\cite{vapnik1999overview} that makes prediction by considering relation between instances in the test set, Liu \etal~\cite{liu2018learning} propose transductive propagation network (TPN) that learns to propagate labels from labeled instances to unlabeled test instances. However, it tries to propagate labels by exploiting the manifold structure of all test samples within several classes, which is insufficient for transductive inference.

In this paper, we propose a novel Progressive Cluster Purification (PCP) method for transductive few-shot learning. Unlike \cite{ren2018meta, liu2018learning}, our method tries to model data relation within a cluster as well as among different clusters, which is more effective for transductive inference. The PCP can progressively purify the cluster of each prototype by exploring the manifold structure of each semantic cluster (shown in Fig.\ref{fig:cover}).  The architecture of the proposed PCP is shown in Fig.\ref{fig:Architecture}. Specifically, the PCP first extracts features for input images, and then performs two-level operations: inter-class classification and intra-class transduction. For the inter-class classification, a classification module is trained to learn a deep distance metric which is used to compare the test samples with prototypes and partition all the test samples into several clusters. Each cluster will gather a lot of instances with the same semantics corresponding to the prototype. Therefore, in each cluster, the intra-class transduction can effectively infer relation between samples by exploiting inherent manifold structure of each cluster. Then, it computes the degree of relation for each instance in a cluster and selects top-$L$ instances with the highest degrees of relation to refine the prototype of this cluster, which can improve the prototype to represent the real distribution of semantic clusters. With these refined prototypes, we remeasure all the test instances to purify each cluster. Furthermore, the inter-class classification and the intra-class transduction are extremely flexible to be repeated several times to progressively purify the clusters.

The main contributions of this work are summarized as follows:
\begin{itemize}
  \item We propose a novel Progressive Cluster Purification (PCP) method for transductive few-shot learning, which can progressively purify the cluster of each prototype by exploring the semantic interdependency in the individual cluster space.
  %explore the fine-grained semantic interdependency in the individual cluster space of each prototype.
  \item We propose an intra-class transduction operation for transductive inference, which can effectively exploit the inherent manifold of each semantic cluster.
  \item The proposed method achieves the state-of-the-art results on both miniImagenet and tieredImageNet dataset. We perform ablation studies to confirm the effectiveness of our model.
\end{itemize}

\section{Related Work}
In this section, we briefly review recent progress on few-shot learning and introduce the idea of transductive inference and its application on few-shot learning.

%\textbf{Meta-learning}
\textbf{\emph{Meta-learning}} \hspace{3mm}
Meta-learning methods aim to extract task-agnostic knowledge from a sequence of similar tasks and apply it to new tasks for fast adaption. The task-agnostic knowledge has flexible
representation, thus leading to different meta-learning methods. Inspired by stochastic gradient descent (SGD)~\cite{bottou2010large}, Ravi \etal~\cite{ravi2016optimization}
employ a LSTM~\cite{hochreiter1997long} network to control parameter optimization process, which leads to better generalization than SGD when training on few labeled samples. MAML~\cite{finn2017model} learns sensitive and general initial parameters, which can fast adapt to a new task with only one or few steps of gradient-descent update. Both of these two methods are categorized to gradient-based approaches, which need backward computation on target loss. Motivated by the correlation between classifier parameters and network activations, Qiao \etal~\cite{qiao2018few} directly predict the parameters of the classifier from the activations. The predictor is learned ene-to-end on many classes in training set. All of these meta-learning methods benefit from the episodic training strategy, \ie training on a series of similar tasks, which we also employ in our method.
%Considering the complexity of hand-designed meta-learners, Mishra \etal~\cite{mishra2017simple} proposed a simple and generic meta-learner architecture (SNAIL), which consists of multiple layers of temporal convolution and soft attention. The temporal convolution layer can aggregate information from past and the soft attention layer can extract the most important pieces.

%\textbf{Metric Learning}
\textbf{\emph{Metric Learning}} \hspace{3mm}
Metric learning methods aim to learn a general embedding space, where samples with the same semantics are close to each other. In this pipeline, Matching networks~\cite{vinyals2016matching} employ a simple cosine distance metric to measure the similarity between two samples and categorize test samples with a KNN classifier. Instead of simple distance metrics, Sung \etal~\cite{sung2018learning} propose to learn a parameterized network to measure the similarity between two samples, which looses the constraint on embedding function and therefore leads to better distance metrics. A simple but effective metric learning method, the prototypical networks~\cite{snell2017prototypical}, extracts a prototype from few labeled samples to represent a new concept. However, due to the scarcity of labeled samples, it is hard to learn a generalized semantic space with the prototype. Ren \etal~\cite{ren2018meta} propose novel extensions of the prototypical networks that use external unlabeled examples to facilitate prototype production. However, a large number of unlabeled samples from different classes will interfere with the production of more trustworthy prototypes. Our method can be categorized into prototype-based methods. Benefitting from exploiting manifold structure of semantic clusters, our method can refine the prototype to better represent the real distribution of semantic clusters.

%\textbf{Sample Synthesis}
\textbf{\emph{Sample Synthesis}} \hspace{3mm}
A straightforward idea to deal with data scarcity is to synthesize extra samples based on given data. In this pipeline, Wang \etal~\cite{wang2018low} propose a hallucinator that takes as input a given instance and noise vectors and generates new instances from different views to train a classifier. They directly train the hallucinator and the classifier end-to-end in a meta-learning manner. Instead of noise vectors, Schwartz \etal~\cite{schwartz2018delta} modify an auto-encoder to extract transferable intra-class deformations (or "deltas"), which are applied to novel samples to synthesize more useful samples. Different from the above approaches that directly use the synthesized samples to train the classifier, Zhang \etal~\cite{zhang2018metagan} train the classifier in a generative adversarial framework, \ie, the classifier can discriminate whether a sample is real or synthesized. In this way, they claim that imperfect
generators provide fake data between the manifolds of different real data classes and therefore lead to much sharper classification decision boundaries. In our method, we also enrich the training data with extra samples, which are from the test set rather than synthesized.

%\textbf{Transductive Inference}
\textbf{\emph{Transductive Inference}} \hspace{3mm}
Transductive inference or transduction is first introduced in ~\cite{vapnik1999overview}, which aims to utilize both training and test set to solve a specific problem, instead of
generating a static model only from the training data.
%Compared to another confusing method, semi-supervised learning, transductive inference focuses on improving the performance on test set by considering the manifold structure of the test samples rather than utilizing unlabeled samples.
An example of transductive inference on classification task is Transductive Support
Vector Machine (TSVM)~\cite{joachims1999transductive}, which iteratively refines decision boundaries to maximize the interval between decision boundaries for both training and test
samples. Utilizing the manifold structure of test set, transductive inference has natural advantages in dealing with the problem of data scarcity, and therefore is introduced for
few-shot learning in recent literature. Nichol \etal ~\cite{DBLP:journals/corr/abs-1803-02999} experiment with a transductive setting and share information among test samples via batch
normalization. Liu \etal~\cite{liu2018learning} first employ a graph neural network (GNN)~\cite{kipf2016semi} to explicitly model transductive inerence as label propagation. However, they try to exploit the manifold structure of all test samples within several classes, which is inefficient for transductive inference. Our intra-class transduction can effectively infer relation between samples by exploiting the inherent manifold structure of each cluster.
%Inspired by transductive inference, in this paper, we are devoted to refining the prototypes by exploring the relationship among the test samples, thus improving the performance on the test set.

%------------------------------------------------------------------------
\section{Methodology}

In this section, we first briefly review the problem definition of few-shot classification task and then introduce our progressive cluster purification method.

\subsection{Problem Definition}

We demonstrate our method on few-shot classification task with episodic paradigm proposed by Vinyals \etal~\cite{vinyals2016matching}. Given a large labeled training set with a set of classes $\mathcal{C}_{train}$, the goal of few-shot learning is to learn to generalize classifiers to novel classes $\mathcal{C}_{test}$ with a few labeled examples. In each training episode, there is a support set and a query set. Under typical $N$-way $K$-shot setting, the support set $\mathcal{S}=\{ (x_1,y_1),(x_2,y_2),...,(x_{N \times K},y_{N \times K})\}$ randomly selects $K$ samples from each of the $N$ classes that are sampled from $\mathcal{C}_{train}$. The query set $\mathcal{Q}=\{ (x_1^*,y_1^*),(x_2^*,y_2^*),...,(x_{N \times M}^*,y_{N \times M}^*)\}$ contains a fraction of the remainder data in these $N$ classes. Based on the support set, we are required to obtain classifiers for the query set with a learnable model, which is optimized by minimizing the prediction loss of the query set. The training process is carried out over many episodes until the model converge. The test set $\mathcal{C}_{test}$ is also split into $N$-way $K$-shot episodes, where the average prediction accuracy of the query sets is used to evaluate the ability of the model to recognize novel classes.

Due to the scarcity of the support samples, it is very difficult to obtain a well-generalized classifier by only training it on the support set.
%Ren \etal~\cite{ren2018meta} propose to use external unlabeled examples to assist to generalize classifiers. However, a large number of unlabeled heterogeneous samples will interfere with the production of more trustworthy classifiers.
Liu \etal~\cite{liu2018learning} propose transductive propagation network (TPN) that learns to propagate labels from labeled instances to unlabeled query instances by exploiting the manifold structure of the query set. However, it tries to exploit the manifold structure of all query samples within several classes, which is inefficient for transductive inference. This inspires us to explore the semantic interdependency in individual cluster space.

\subsection{Progressive Cluster Purification}

In this paper, we propose a novel and general Progressive Cluster Purification (PCP) method for transductive few-shot learnng, which is more effective to exploit the manifold structure of the query set. The whole framework of PCP is illustrated in Fig.~\ref{fig:Architecture}. In PCP, a feature extraction network first extracts discriminative visual features from input images. Then an inter-class classification module partitions the query set into several clusters. Next, an intra-class transduction module infers relation between samples by exploiting inherent manifold structure of each cluster and refines the prototype to better represent the real distribution of its corresponding cluster. With these refined prototypes, we remeasure all the query instances to purify each cluster.

%To better exploit the manifold structure of the test set for transductive few-shot learning, we propose a Progressively Cluster Purifying (PCP) method for transductive few-shot learnng. As illustrated in Fig.~\ref{fig:Architecture}, it consists of four key parts: a feature extraction network to extract discriminative visual features for input images, an
%inter-class classification module to partition the query set into several clusters with prototypes, an intra-class transduction module to infer relation between samples by
%exploiting inherent manifold structure of each cluster and a prototype refine module to improve the ability of the prototype to represent the intra-class variability. We summarize
%our method with Alg.~\ref{alg:1}.

%\textbf{Initialization}
\textbf{\emph{Feature Extraction \& Prototype Production}} \hspace{3mm}
Given a sample $x_i$ in the support set $\mathcal{S}$ and a sample $x_i^*$ in the query set $\mathcal{Q}$, we employ a convolutional neural network (CNN)~\cite{lecun1989backpropagation} $f_{\phi}$ to extract visual features $f_{\phi}(x_i)$ and $f_{\phi}(x_i^*)$. After that, the initial prototypes $\mathcal{P}^0 = \left\{p_n^0\right\}|_{n=1}^N$ are produced by averaging the feature maps of support samples in the same class:
\begin{equation}
p_{n}^0=\frac{1}{K}\sum_{y_{i}=n}f_{\phi}(x_{i})
\label{equ:1}
\end{equation}
where $p_n^0$ denotes the initial prototype of the $n^{th}$ class, $y_i$ is the label of the support sample $x_i$. Due to scarce labeled data, it is hard for the initial prototypes to represent the real distribution of semantic clusters. Therefore, we consider refining the initial prototypes with transductive inference by inter-class classification and intra-class transduction.

%We average the feature maps of the
%support samples from the same class to construct the initial prototypes $\left\{p_n^0\right\}|_{n=1}^N$, illustrated as:
%\begin{equation}
%p_{n}^0=\frac{1}{K}\sum_{y_{i}=n}f_{\phi}(x_{i})
%\label{equ:1}
%\end{equation}
%During the following refinement iterations,we will iteratively get better prototypes $p_n^1, p_n^2,...,p_n^t$. For the sake of simplicity, we remove the superscript $t$ of all variables in the following equations if it is not necessary.

\begin{figure}[t]
\begin{center}
\includegraphics[width=1.0\linewidth]{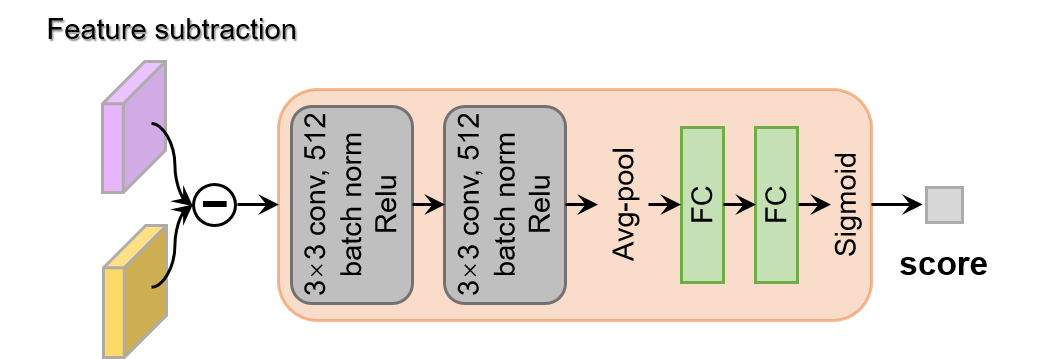}
\end{center}
   \caption{The network structure of the classification module.}
\label{fig:2}
\end{figure}

\textbf{\emph{Inter-class Classification}} \hspace{3mm}
In order to recognize the query samples,
%Unlike \cite{ren2018meta, liu2018learning}, we propose to exploit a fine manifold structure of each cluster for transductive inference. To partition the query samples,
we propose a classification module to measure the semantic similarity between a query sample and a prototype. The detailed structure of classification module is shown in Fig.\ref{fig:2}. Given a prototype $p_{n}^t$ and a query sample $x_{i}^*$, we first compute their feature difference $|f_{\phi}(x_{i}^*)-p_{n}^t|$ with subtraction, where $p_{n}^t$ denotes the initial prototype ($t=0$) or the refined prototype ($t>0$). To learn a deep distance metric between query samples and prototypes, we feed the feature difference $|f_{\phi}(x_{i}^*)-p_{n}^t|$ into a network to adaptively measure the similarity score.

More specifically, the classification network is comprised of two convolutional layers and two fully-connected layers. Each convolutional layer is followed by batch normalization and ReLU. The two fully-connected layers finally regress to the similarity scores between query samples and prototypes. The formulation is defined as follows:
\begin{equation}
 s_{i,n}=\sigma(F(f_{\phi}(x_{i}^*)-p_{n}^t|))
\end{equation}
where $s_{i,n}$ is the similarity score between the query sample $x_{i}^*$ and prototype $p_{n}^t$, $\sigma$ is sigmoid function to map the regression output to the range of [0,1], and $F(\cdot)$ denotes the classification network.

After calculating all similarity scores of query samples with each prototype, we can partition all the query samples into several clusters as follows:
\begin{equation}
\hat{y}_{i}=\mathop{argmax}\limits_{n \in \{1,..,N\}} s_{i,n}
\label{equ:5}
\end{equation}
where $\hat{y}_{i}$ is the predicted label of the query sample $x_{i}^*$.

\begin{figure}[t]
\begin{center}
\includegraphics[width=0.95\linewidth, height=0.60\linewidth]{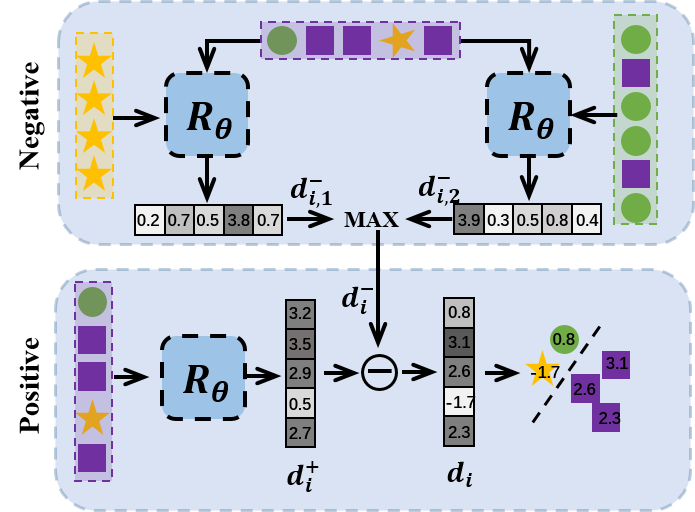}
\end{center}
   \caption{Illustration of the intra-class transduction. The  positive degree $d_i^+$ of a sample represents the semantic consistency within its cluster and the negative degree $d_i^-$ stands for the semantic similarity with other clusters}
\label{translation}
\end{figure}

\textbf{\emph{Intra-class Transduction}} \hspace{3mm}
After inter-class classification, we hope the samples in each cluster would have the same semantics corresponding to the prototype. However, due to the scarcity of the support samples, there are still a small amount samples with different semantics in each cluster. To purify the clusters, transductive inference strategy is employed to deal with scarce-data problem in this paper. Unlike \cite{ren2018meta, liu2018learning}, we propose a intra-class transduction to infer relation between samples by exploiting inherent manifold structure of each cluster. The intra-class transduction is illustrated in Fig.\ref{translation}.

Specifically, we explore pair-wise relation between samples. Due to gathering a lot of samples with the same semantics in a cluster, the relation scores of a sample with other sampeles are aggregated to serve as its degree of relation that indicates its probability to be clustered correctly. In a cluster, the samples with different semantics may be mismeasured with high degree of relation. To weaken this trend and make the degree more trustworthy, the relation of a sample is considered not only within its cluster but also among other clusters. Accordingly, the degree $d_i$ of a sample is computed with a positive degree $d_i^+$ and a negative degree $d_i^-$ (shown in Fig.\ref{translation}). The positive degree of a sample represents the semantic consistency within its cluster and the negative degree stands for the semantic similarity with other clusters.

To calculate the degree of relation for each sample, we employ a relation module to measure the relation score between samples. Note that the network structure of relation module is same as the inter-class classification module. The relation score $r_{i,j}$ of query sample $x_{i}^*$ and $x_{j}^*$ can be calculated as:
\begin{align}
r_{i,j} = \sigma(R_{\theta}(|f_\phi(x_i^*)-f_\phi(x_j^*)|))
\end{align}
where $R_{\theta}$ denotes the relation module. For the query sample $x_i^*$, we average its relation scores with other query samples within its cluster to serve as the positive degree. The formulation is defined as follows:
\begin{align}
d_i^+ &= \frac{1}{N_{\hat{y}_i}}\sum_{\hat{y}_j=\hat{y}_i}r_{i,j} %\nonumber\\
%&= \frac{1}{N_{\hat{y}_i}}\sum_{\hat{y}_j=\hat{y}_i} \sigma(R_{\theta}(|f_\phi(x_i^*)-f_\phi(x_j^*)|))
\label{positive_degree}
\end{align}
where $N_{\hat{y}_i}$ is the number of query samples in the cluster of $x_{i}^*$.
%prototype $p_{\hat{y}_i}^t$. $\hat{y}_i, \hat{y}_j$ denotes the predicted labels of query samples $x_i^*, x_j^*$ that are in the cluster of prototype $p_{\hat{y}_j}^t$.
%The higher the positive degree, the stronger relation to other samples in its cluster, the more likely it is clustered correctly.

The negative degree aims to reduce the confidence of misclassified sample in a cluster. For the query sample $x_i^*$, we first calculate its degree to other clusters:
\begin{align}
d_{i,n}^- &= \frac{1}{N_{n}}\sum_{\hat{y}_j=n} r_{i,j}, \hspace{3mm} n \ne \hat{y}_i %\nonumber\\
%&= \frac{1}{N_{n}}\sum_{\hat{y}_j=n} \sigma(R_{\theta}(|f_\phi(x_i^*)-f_\phi(x_j^{*n})|)), \hspace{0.5mm}  \hat{y}_i \ne n
\label{equ:8}
\end{align}
where $N_n$ is the number of query samples in the cluster of prototype $p_{n}^t$, and $d_{i,n}^-$ is the degree of relation between $x_i^*$ and this cluster. Based on the inherent manifold structure of a semantic cluster, we can infer that higher degree of relation should exist between a misclassified sample and its ground-truth cluster. Therefore, we select the maximum degree to other cluster as the negative degree, which can be formulated as:
\begin{equation}
d_i^-=\mathop{max}\limits_{n\neq\hat{y}_i} d_{i,n}^-
\end{equation}
To balance these two degrees, we fuse them with a weight $\lambda$ and calculate the final degree of $x_i^*$ as:
\begin{equation}
d_i=d_i^+-\lambda d_i^-
\label{degree}
\end{equation}

After calculating the degree of each sample, we sort the query samples in each cluster according to their degrees and select top-$L$ samples to refine the corresponding prototype:
\begin{equation}
p_n^{t}=\frac{1}{L+1}(p_n^{t-1}+\sum_{l=1}^{L}f_{\phi}(x_{nl}^*))\\
\end{equation}
where $p_n^{t}$ is refined prototype at $t^{th}$ time, $x_{nl}^*$ is the query sample with the $l^{th}$ highest degree in the cluster of $p_n^{t-1}$. After refinement, the prototypes can better represent the real distribution of semantic clusters

%%\textbf{\emph{Prototype Refinement}} \hspace{3mm}
%After intra-class transduction, we get more trustworthy prediction for each query sample, with which we can improve the ability of the prototype to represent the intra-class
%variability. Specifically, we propose a hard refinement strategy that we sort the query samples in each cluster according to their degrees and only keep top-$L$ samples with the highest
%degrees. We enrich the support set with these new samples and correspondingly refine the prototype as:
%\begin{equation}
%p_n^{t}=\frac{1}{L+1}(p_n^{t-1}+\sum_{i=1}^{L}f_{\phi}(q_{n,i}))\\
%\end{equation}
%where $q_{n,i}$ is the query sample with the $i^{th}$ highest score in the $n^{th}$ cluster.

\textbf{\emph{Iterative Refinement and Final Prediction}} \hspace{3mm}
With these refined prototypes, we can remeasure all the test samples to purify each cluster. Meanwhile, there will be more compact manifold structure of each cluster, so we can continue our prototype refinement iteratively. After $T$ iterations, we will get converged clusters and accordingly obtain final classification results for the query samples. The iterative process is illustrated with Alg.~\ref{alg:1}

\begin{algorithm}
\caption{Progressive Cluster Purification for Transductive Few-shot Learning.}
\KwIn{$\mathcal{S} = \left\{(x_1, y_1),...,(x_{N \times K}, y_{N \times K})\right\}$, $\mathcal{Q} = \left\{x_1^*,...,x_{N \times M}^*\right\}$ }
$\bullet$ Feature extraction with $f_{\phi}$ and prototype production: $p_n^0\leftarrow\frac{1}{K}\sum_{y_{i}=n}f_{\phi}(x_{i})$ \;

%\KwOut{$\hat{y}_1^T, \hat{y}_2^T,...,\hat{y}_M^T$}

\For{$t=0; t<T; t=t+1$}
{
    \textbf{\emph{\# Iterative Refinement \; }}
    $\bullet$ Inter-class classification partitions $\mathcal{Q}$ into $N$ clusters with prototypes $\{p_1^t,...,p_N^t\}$ \;
    \For{$n=1; n\le N; n=n+1$}
    {
        $\bullet$ Intra-class transduction calculates the final degree $d_i$ of a sample $x_i^*$ in this cluster\;
        $\bullet$ Sort the query samples and select top-$L$ samples with highest degrees of relation \;
        $\bullet$ Refine the prototype of this cluster: $p_n^{t+1}\leftarrow \frac{1}{L+1}(p_n^{t}+\sum_{l=1}^{L}f_{\phi}(x_{nl}^*))$

    }
}
$\bullet$ Inter-class classification partitions $\mathcal{Q}$ into $N$ clusters with prototypes $\{p_1^T,...,p_N^T\}$ \;
%Inter-class classification predicts the label $\hat{y}_i^t$ for each $x_i^*$ to partition $\mathcal{Q}$ into $N$ clusters\;
\KwOut{the predicted label $\hat{y}_1, \hat{y}_2,...,\hat{y}_{N \times M}$}
\label{alg:1}
\end{algorithm}

%\subsection{Learning PCP}
%For the inter-class classification module, we supervise it by minimize the mean squared error to regress the similarity score:
%\begin{equation}
%\small
%\mathcal{L}_{inter}=\frac{1}{M \times N} \sum_{i}^{M \times N} \frac{1}{ N} ( (1-s_{i,y_{i}} )^2+\sum_{n=1,n\neq y_{i}}^{N} s_{i,n}^2 )
%\label{equ:12}
%\end{equation}
%where $y_i$ is the ground truth label of query sample $x_i^*$.
%%With this loss, we expect that the similarity between a query sample and its ground truth prototype will approach one and the its
%%similarity with other prototypes will approach zero.
%For the intra-class transduction, the loss function is calculated as:
%\begin{equation}
%\mathcal{L}_{intra}(q_i,q_j)=
%\begin{cases}
%(1-s_{i,j})^2&,y_i=y_j\\
%s_{i,j}^2&,y_i\neq y_j
%\end{cases}
%\label{equ:13}
%\end{equation}

\subsection{Learning PCP}
For the inter-class classification module, we optimize it by minimize the mean squared error of the regressed similarity score between each query sample and each prototype:
\begin{equation}
\small
\mathcal{L}_{inter}=\frac{1}{N\times M} \sum_{i}^{N \times M} ( (1-s_{i,y_{i}} )^2+\sum_{n=1,n\neq y_{i}}^{N} s_{i,n}^2 )
\label{equ:12}
\end{equation}
where $y_i$ is the ground truth label of the query sample $x_i^*$.
%With this loss, we expect that the similarity between a query sample and its ground truth prototype will approach one and the its
%similarity with other prototypes will approach zero.
For the intra-class transduction, the loss function is calculated as:
\begin{equation}
\mathcal{L}_{intra}(q_i,q_j)=
\begin{cases}
(1-r_{i,j})^2&,y_i=y_j\\
r_{i,j}^2&,y_i\neq y_j
\end{cases}
\label{equ:13}
\end{equation}

\begin{equation}
\mathcal{L}_{intra}=\frac{1}{(N\times M)^2}\sum_{i=1}^{N\times M}\sum_{j=1,j\neq i}^{N \times M}\mathcal{L}_{intra}(q_i,q_j)
\label{equ:14}
\end{equation}

Considering that all modules are shared among iterations, we train our model with only one iteration and test it with more iterations. Specifically, we first train the feature extractor and the inter-class classification module with the loss Eqn.~\ref{equ:12}, and then fix them to train the relation module for intra-class transduction with the loss Eqn.~\ref{equ:14}.

%------------------------------------------------------------------------
\section{Experiments}

In our experiments, we evaluate the proposed PCP on two few-shot learning datasets: miniImageNet dataset and tieredImageNet dataset. The analysis of experimental results confirms the effectiveness of our model for few-shot learning.

\subsection{Datasets}

\textbf{MiniImageNet Dataset.}
This dataset is a subset of the large dataset ILSVRC-12~\cite{ILSVRC15}, which is first proposed in ~\cite{vinyals2016matching} as a benchmark for few-shot learning. It contains 100 classes and 600 images in each class. All images are resized to $84\times84$ pixels. We follow the splits proposed by Ravi~\etal~\cite{ravi2016optimization}: 64 classes for training, 16 classes for validation and 20 classes for testing.

%Under the typical $N$-way $K$-shot setting, for a test episode, we randomly select $N$ classes from 20 test
%classes and $K$ samples from each selected class to constitute a support set. We select another $M$ samples from each of the selected classes to constitute a query set. We average the
%accuracy on query sets over 10,000 episodes with 95\% confidence interval to get more accurate evaluation. Note that we only use the validation set to early stop training in
%order to reduce overfitting.

\textbf{TieredImageNet Dataset.}
This dataset is also a subset of ILSVRC-12~\cite{ILSVRC15} but is larger than miniImageNet. It contains 608 classes, which can be grouped into 34 broader categories according to
high-level semantics. There are 10$\sim$30 classes in each category. The average number of images in each class is 1281. All the images are also resized to $84\times84$ pixels. Following the splits in ~\cite{ren2018meta}, we use 20 categories for training (351 classes), 6 for validation (97 classes) and 8 for testing (160 classes). Note that there are large semantic differences between categories, which ensures that the training classes are distinct from the test classes. Therefore, this is a more challenging and realistic dataset for few-shot learning.

%Note that the semantic difference between categories are much greater than the semantic difference between classes. Therefore, this is a
%more challenging dataset and suits realistic few-shot scenario better. The constitution of episodes for this dataset is the same as miniImageNet and we also use validation set only
%for early stop.

\subsection{Implementation Details}

To farely compare with other methods~\cite{snell2017prototypical,liu2018learning,sung2018learning}, we employ the same four-layer convolution network (ConvNet) as the feature extraction module. More concretely, each $3\times3$ convolutional layer is followed by batch normalization and ReLU, and $2\times2$ max-pooling layers are applied to reduce the size of feature maps. To constitute an episode, we select $M$=15 query samples for each class. Similar to \cite{liu2018learning}, a ``Higher Shot'' training strategy is adopted during training, which uses more sample to serve as the support set (larger $K$).  In our experiments, we employ 5-shot and 10-shot to train the model for 1-shot and 5-shot test task, respectively. For intra-class transduction, we select $L$=9 query samples to refine the prototype. We set $\lambda$=0.8 and 0.6 for miniImageNet dataset and tieredImageNet dataset, respectively. During training, we use the Adam optimizer \cite{kingma2014adam} to optimize our model. The initial learning rate is set to 0.001. During testing, prediction accuracies of query sets are averaged over 10,000 episodes with 95\% confidence interval.

%accuracy on query sets over 10,000 episodes with 95\% confidence interval to get more accurate evaluation. Note that we only use the validation set to early stop training in
%order to reduce overfitting.

%To farely compare with other methods~\cite{snell2017prototypical,liu2018learning,sung2018learning}, we employ the same 4-layer network (ConvNet) to extract visual features. Each layer of ConvNet contains $3\times3$ convolution, batch normalization and ReLU. $2\times2$ max pooling is applied to reduce the size of feature maps. To constitute an episode, we select $M$=15 query samples for each class. Follow ~\cite{liu2018learning}, we also employ high-shot trick for training, \ie, we use more support samples (larger $K$) to train the model but test it under standard settings. For intra-class transduction, we select $L$=9 query samples to refine the prototype. We set $\lambda$=0.6 to balance the positive degree and the negative degree. Our main results are based on 3 iterations of refinement. During training, we adopt Adam~\cite{kingma2014adam} optimizer with initial learning rate as 0.001.

\subsection{Experimental Results}
In this section, we compare our proposed PCP with several state-of-the-art methods on the used two datasets.

\subsubsection{MiniImageNet Dataset}

\begin{table}[htbp]
    \begin{center}
    \begin{tabular}{l|cc}
        \toprule[1px]
        Methods&5-way 1-shot&5-way 5-shot\\
        \hline
        \hline
        MatchingNet~\cite{vinyals2016matching}&43.56$\pm$0.84&55.31$\pm$0.73\\
        MAML~\cite{finn2017model}&48.70$\pm$1.84&63.11$\pm$0.92\\
        ProtoNet~\cite{snell2017prototypical}&49.42$\pm$0.78&68.20$\pm$0.66\\
        RelationNet~\cite{sung2018learning}&50.44$\pm$0.82&65.32$\pm$0.70\\
        MM-Net~\cite{cai2018memory}&53.37$\pm$0.48&66.97$\pm$0.35\\
        Qiao \etal~\cite{qiao2018few}&54.53$\pm$0.40&67.87$\pm$0.20\\
        TPN~\cite{liu2018learning}&55.51$\pm$*&69.86$\pm$*\\
%        \cline{2-4}
        \hline
        Ours&\textbf{58.40$\pm$0.27}&\textbf{70.66$\pm$0.19}\\
%        \hline
%        \multirow{6}{*}{ResNet}&SNAIL~\cite{}&55.71$\pm$0.99&68.88$\pm$0.92\\
%        &TADAM~\cite{}&58.50$\pm$0.30&76.70$\pm$0.30\\
%        &TPN~\cite{}&59.46$\pm$*&75.65$\pm$*\\
%        &Qiao \etal~\cite{}&59.60$\pm$0.41&73.74$\pm$0.19\\
%        &Leo~\cite{}&61.76$\pm$0.08&77.59$\pm$0.12\\
%        \cline{2-4}
%        &Ours(ResNet)&&\\
        \bottomrule[1px]
    \end{tabular}
    \end{center}
    \caption{Test accuracy of the state-of-the-art methods on miniImageNet dataset under 5-way 1-shot and 5-way 5-shot settings. All results are given based on the same 4-layer ConvNet with 95\% confidence intervals. "*" means the confidence interval is not given by the author.}
    \label{tab:miniImageNet}
\end{table}

The experimental results on miniImageNet dataset are shown in Tab.~\ref{tab:miniImageNet}. we can see that our proposed method achieves the best performance of 58.40\% and 70.66\% for 1-shot and 5-shot settings, respectively. To demonstrate the effectiveness of our method, we choose the following related methods to compare and analyze the results:

ProtoNet~\cite{snell2017prototypical} and RelationNet~\cite{sung2018learning} are proposed to average/sum the features of a support set as a prototype to represent a new class. On the 1-shot setting, the results of ProtoNet and RelationNet are 49.42\% and 50.44\%. Compared with them, the proposed PCP uses the unlabled test samples to refine the prototypes with transduction strategy. The PCP achieves the accuracy of 58.40\%, which outperforms 8.98\% and 7.96\% than ProtoNet and RelationNet, respectively. The comparison results prove that our refined prototypes are more effective to represent the distribution of query set.
%ProtoNet~\cite{snell2017prototypical} and RelationNet~\cite{sung2018learning} also employ prototypes to represent new concepts. However, due to the scarcity of support samples, the prototype can not represent the whole semantic space. From Tab.~\ref{tab:miniImageNet}, we can see that our method outperforms them by 7.93\% and 2.46\% under 1-shot and 5-shot settings
%respectively, which proves that our prototype refinement is effective to improve the ability of the prototype to represent the intra-class variability.
%Qiao \etal~\cite{qiao2018few} propose to directly predict classifier parameters from activations for few-shot learning, which can be seen as an induction method. In contrast, utilizing the manifold structure of test samples, our transduction based method outperforms theirs by 3.84\% and 2.79\%  under 1-shot and 5-shot settings.
Another related work is TPN~\cite{liu2018learning} that proposes to propagate labels from labeled instances to unlabeled test instances with a transductive propagation network. Although TPN employs transductive inference for few-shot learning, it does not explicitly exploit manifold structure in a semantic cluster. The intra-class transduction in the PCP exploits inherent manifold structure of each cluster. Our performances outperform the TPN by 2.98\% and 0.8\% for 1-shot and 5-shot settings respectively, which confirm our method is more efficient for transduction.

Although our PCP outperforms these advanced methods for 1-shot and 5-shot settings, we can observe that the improved performances of 5-shot are less than that of 1-shot setting. The reason is that the advantage of transduction will be decreased with the increase of training data, which has been proved in \cite{joachims1999transductive}.

%TPN~\cite{liu2018learning} also employs transductive inference for few-shot learning and models it as label propagation with a graph neural network. However, it does not explicitly exploit manifold structure in a semantic cluster, which may be inefficient for transductive inference. In contrast, our intra-class transduction can efficiently infer relation between samples by exploiting inherent manifold structure of each cluster. The experimental results in Tab.~\ref{tab:miniImageNet} show that our method outperforms TPN under both 1-shot and 5-shot settings by 2.86\% and 0.8\% respectively, which confirms our method is more efficient for transduction.
\subsubsection{TieredImageNet Dataset}

\begin{table}[htbp]
    \begin{center}
    \begin{tabular}{l|cc}
        \toprule[1px]
        Methods&5-way 1-shot&5-way 5-shot\\
        \hline
        \hline
        Reptile(from ~\cite{liu2018learning})&48.97&66.47\\
        MAML(from ~\cite{liu2018learning})&51.67&70.30\\
        ProtoNet(from ~\cite{liu2018learning})&53.31&72.69\\
        RelationNet(from ~\cite{liu2018learning})&54.48&71.31\\
        TPN~\cite{liu2018learning}&59.91&73.30\\
        \hline
        Ours&\textbf{62.43}&\textbf{74.11}\\
        \bottomrule[1px]
    \end{tabular}
    \end{center}
    \caption{Test accuracy of the state-of-the-art methods on tiredImageNet dataset under 5-way 1-shot and 5-way 5-shot settings. For fair comparison, all methods use the same 4-layer ConvNet for feature extraction.}
    \label{tab:tieredImageNet}
\end{table}

The experimental results on tieredImageNet dataset are shown in Tab.~\ref{tab:miniImageNet}. The proposed PCP again achieves the best accuracy of 62.43\% and 74.11\% for 1-shot and 5-shot settings. Compared with ProtoNet~\cite{snell2017prototypical} and RelationNet~\cite{sung2018learning}, our PCP achieves at least 7.95\% and 1.42\% improvement of accuracy for 1-shot and 5-shot respectively. Moreover, our method outperforms TPN~\cite{liu2018learning} by 2.52\% and 0.81\% under 1-shot and 5-shot respectively. The comparing results again confirm the effectiveness of our progressively cluster purifying method for few-shot learning.

%Although the semantic difference between training set and test set is very large in tieredImageNet, we have still achieved the best performance on this dataset. From
%Tab.~\ref{tab:tieredImageNet}, we can see that our method outperforms TPN~\cite{liu2018learning} by 2.52\% and 0.81\% under 1-shot and 5-shot respectively, which shows that our method
%can generalize better even with huge semantic gap. Compared with ProtoNet~\cite{snell2017prototypical} and RelationNet~\cite{sung2018learning}, we achieve at least 7.95\% and 1.42\%
%improvement of accuracy under 1-shot and 5-shot respectively, which again confirms the effectiveness of our progressively cluster purifying method for few-shot learning.

\subsection{Model Analysis}

In this section, we first analyze the proposed method by ablating it into several baselines.
%To understand the properties of the proposed PCP, we first analyze the effectiveness of several key components on both miniImageNet dataset and tieredImageNet dataset.
Then, we study the influences of several parameters, \ie the number of refinement iterations $T$, the number of selected samples $L$ for hard refinement strategy and the weight coefficient $\lambda$  for balancing the positive degree and the negative degree.

%In this section, we analyze the proposed method by ablating it into several baselines. We also study the effect of number of refinement iterations on test accuracy, which shows that our method can rapidly reach convergence. Then, we experiment with different numbers of selected samples for our hard refinement strategy. Finally, we study how to choose a suitable $\lambda$ to balance the positive degree and the negative degree.

\subsubsection{Ablation Study}

\begin{table}[htbp]
    \centering
    \begin{tabular}{l|cc|cc}
        \toprule[1px]
        \multirow{2}{*}{Methods}&\multicolumn{2}{|c|}{miniImageNet}&\multicolumn{2}{c}{tieredImageNet}\\
        &1-shot&5-shot&1-shot&5-shot\\
        \hline
        \hline
        Baseline          &51.77&66.68&53.95&69.64\\
        Baseline\emph{+ref}       &55.39&68.28&57.60&70.86\\
        Baseline\emph{+sel}       &55.72&68.72&58.21&71.56\\
        Baseline\emph{+intra$^+$}    &56.58&69.77&59.18&72.45\\
        Ours               &57.18&70.58&59.82&73.00\\
        Ours(high-shot)&\textbf{58.37}&\textbf{70.66}&\textbf{62.43}&\textbf{74.11}\\
        \bottomrule[1px]
    \end{tabular}
    \caption{Ablation study results of our method on miniImageNet and tieredImageNet datasets under 5-way 1-shot and 5-way 5-shot settings.}
    \label{tab:3}
\end{table}

To validate the effectiveness of out PCP in few-shot learning, we conduct experiments with different configurations on miniImageNet dataset and tieredImageNet dataset as follows:

\textbf{Baseline}\hspace{1mm} For this model, a feature extraction network extracts image features and produces prototypes, and an inter-class classification classifies the query images. It is similar to ~\cite{sung2018learning}.

%The baseline model consists of a feature extraction network and inter-class classification

\textbf{Baseline\emph{+ref}}\hspace{1mm} Compare with Baseline, this model averages all the features in a cluster to refine the corresponding prototype. the refined prototypes are used to recognize the query images.

\textbf{Baseline\emph{+sel}}\hspace{1mm} Compare with averaging all the features in Baseline\emph{+ref}, this model will select top-$L$ samples according to the similarity between the query sample and the prototype to refine the corresponding prototype.

\textbf{Baseline\emph{+intra$^+$}}\hspace{1mm} Compare with Baseline\emph{+sel}, this model applies the intra-class transduction to refine the prototypes by exploit inherent manifold structure of each cluster. However, the intra-class transduction only use the positive degree (Eqn.\ref{positive_degree}) to select samples.

\textbf{Ours}\hspace{1mm} This denotes our proposed model that considers the positive degree and the negative degree to select samples (shown as Eqn.\ref{positive_degree}$\sim$\ref{degree}).

\textbf{Ours(high-shot)} Similar to \cite{liu2018learning}, a ``Higher Shot'' training strategy is adopted during training.

Tab.~\ref{tab:3} shows the comparison results of the variants and our proposed model. Compared with Baseline, Baseline\emph{+ref} and Baseline\emph{+sel} achieve much higher accuracies, which illustrates the advantage of transduction for few-shot learning. With the intra-class transduction to exploit inherent manifold structure of each cluster, Baseline\emph{+intra$^+$} can further improve the performances. Compared with Baseline\emph{+intra$^+$}, our PHP considers the positive degree and the negative degree simultaneously to reduce the confidence of misclassified sample in the cluster. The improved results confirm our method is effective to select more trustworthy samples to refine the prototype. Similar to \cite{liu2018learning},``Higher Shot'' can improve the performances of PCP£¬which confirms that our PCH can effectively exploit inherent manifold structure of each cluster.

\subsubsection{Influence of Parameters}
%\subsubsection{Influence of Purification Iterations}

\begin{table}[b]
    \centering
    \begin{tabular}{c|cc|cc}
        \toprule[1px]
        \multirow{2}{*}{Iterations}&\multicolumn{2}{|c|}{\textbf{miniImageNet}}&\multicolumn{2}{c}{\textbf{tieredImageNet}}\\
        &1-shot&5-shot&1-shot&5-shot\\
        \hline
        \hline
        0&51.77&66.68&54.94&70.40\\
        1&56.72&69.97&60.84&73.36\\
        2&58.00&70.53&62.08&73.76\\
        3&\textbf{58.37}&\textbf{70.66}&\textbf{62.43}&\textbf{74.11}\\
        \bottomrule[1px]
    \end{tabular}
    \caption{The comparison results with different purification iterations on miniImageNet and tieredImageNet datasets under 5-way 1-shot and 5-way 5-shot settings.
    %Test accuracy vs the number of purification iterations on miniImageNet and tieredImageNet datasets under 5-way 1-shot and 5-way 5-shot settings.
    }
    \label{tab:Iterations}
\end{table}

%\begin{figure}[t]
%\begin{center}
%\includegraphics[width=1.0\linewidth]{./fig/Iterations.pdf}
%\end{center}
%   \caption{Test accuracy vs the number of iterations on miniImageNet and tieredImageNet.}
%\label{fig:Iterations}
%\end{figure}

\emph{\textbf{Influence of purification iterations}}\hspace{3mm} To further understand the properties of progressive cluster purification, we analyze the influence of the number of purification iterations. The comparison results are shown in Tab.~\ref{tab:Iterations}. Compared with no refinement, we can observe that the performance is significantly improved with one refinement iteration. It confirms the efficiency of our intra-class transduction operation for transduction. Moreover, we can see that the accuracy gets steadily improved during next few iterations, which illustrates that the model can further enhance the ability of refined prototypes to represent the distribution of semantic cluster.

% and then saturated with the increase of $T$.
%
%the performance increases by a small amount when increasing T, and saturates soon.
%
%
%With our progressive cluster purifying, we can obtain better prediction for test set after each iteration. We plot the accuracy improvement with the number of iterations on miniImageNet and tieredImageNet under 5-way 1-shot and 5-way 5-shot settings, as shown in Tab.~\ref{tab:Iterations}. Compared with no refinement, there are great accuracy improvements even if we employ only one refinement iteration (4.95\% and 3.29\% on miniImageNet under 1-shot and 5-shot respectively, 5.90\% and 2.96\% on tieredImageNet under 1-shot and 5-shot respectively). It confirms the efficiency of our intra-class transduction operation for transduction. From Tab.~\ref{tab:Iterations} we can see that the accuracy gets steadily improved during next
%few iterations, which illustrates that the prototypes can still get further refined with our progressive cluster purifying. We observe that the accuracy stop increasing after 3
%iterations, which is because the prototypes are converged. So we employ 3 iterations in most of our experiments.

%\subsubsection{Hard Refinement Strategy}

\begin{figure}[t]
\begin{center}
\includegraphics[width=1.0\linewidth]{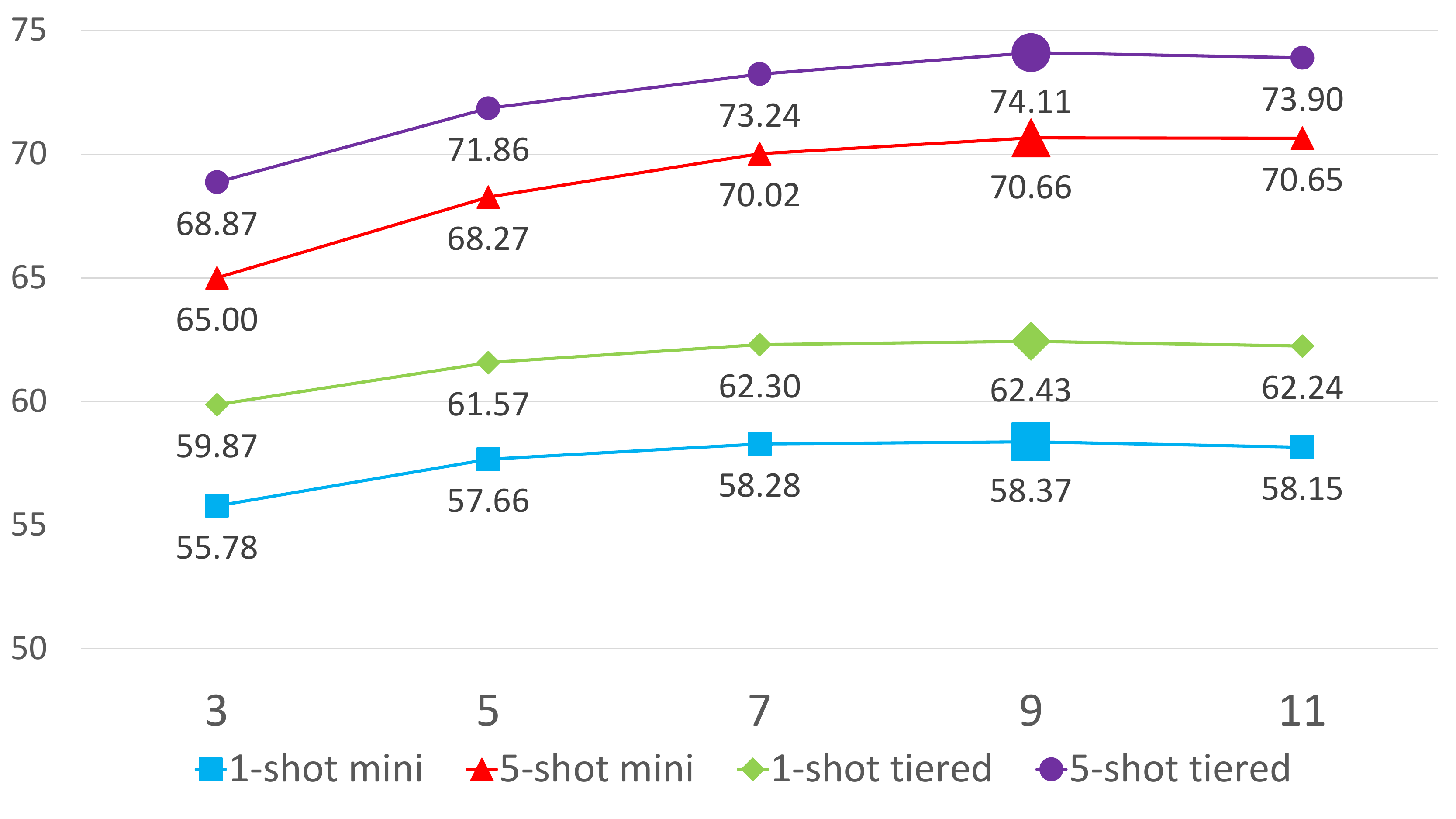}
\end{center}
   \caption{The comparison results with the different number of selected samples $L$ on miniImageNet and tieredImageNet datasets under 5-way 1-shot and 5-way 5-shot settings.
   %Test accuracy  the number of selected samples $L$ on miniImageNet and tieredImageNet datasets under 5-way 1-shot and 5-way 5-shot settings.
   }
\label{fig:L}
\end{figure}

\emph{\textbf{Influence of the number of selected samples}}\hspace{3mm} In intra-class transduction, we sort the query samples in each cluster according to their degrees and select top-$L$ samples to refine the corresponding prototype. We also analyse the influence of the number of selected samples on performance. Fig.~\ref{fig:L} shows the experiment results. With the increase of $L$, the performance is significantly improved and then decreases. The reason is that the selected query samples are almost correctly classified with small $L$. With more trustworthy query samples to refine the prototype, it can better represent the real distribution of semantic cluster. As $L$ gets larger, however, more misclassified samples will be selected to refine prototypes, which will harm the quality of prototypes and result in accuracy drop.

%For our hard refinement strategy, we need to select top-$L$ query samples with the highest degrees in each cluster to refine the prototype. We observe that $L$ has strong effect on
%classification accuracy of test set, so we set a series of $L$ to analyze their relationship. Experiment results on both miniImageNet and tieredImageNet dataset under 5-way 1-shot and 5-way 5-shot settings are shown in Fig.~\ref{fig:L}. It is shown that the test accuracy grows rapidly with $L$ when $L$ is small. This is because the selected query samples are almost correctly
%classified when $L$ is very small. With more trustworthy query samples to refine the prototype, it can better represent the real distribution of semantic cluster, and therefore leads to more accurate prediction for test smaples. As $L$ gets larger and larger, however, more misclassified samples will be selected to refine prototypes, which will harm the quality of prototypes and result in accuracy drop. From Fig.~\ref{fig:L}, we can see that $K=9$ is the best for both miniImageNet and tieredImageNet under both 1-shot and 5-shot settings.

%\subsubsection{Ratio of Negative Degree to Positive Degree}

\begin{table}[htbp]
    \centering
    \begin{tabular}{c|cc|cc}
        \toprule[1px]
        \multirow{2}{*}{$\lambda$}&\multicolumn{2}{|c|}{miniImageNet}&\multicolumn{2}{c}{tieredImageNet}\\
        &1-shot&5-shot&1-shot&5-shot\\
        \hline
        \hline
        0&57.61&69.70&61.43&72.77\\
        0.2&58.04&70.15&62.08&73.55\\
        0.4&58.26&70.47&62.36&73.74\\
        0.6&58.37&\textbf{70.66}&\textbf{62.43}&\textbf{74.11}\\
        0.8&\textbf{58.40}&\textbf{70.66}&62.39&73.73\\
        1.0&58.31&70.46&62.26&73.58\\
        \bottomrule[1px]
    \end{tabular}
    \caption{The comparison results with different weight coefficients on miniImageNet and tieredImageNet datasets under 5-way 1-shot and 5-way 5-shot settings.
    %Test accuracy vs the ratio of the negative score to the positive score on miniImageNet and tieredImageNet datasets under 5-way 1-shot and 5-way 5-shot settings.
    }
    \label{tab:lambda}
\end{table}

\emph{\textbf{Influence of the weight coefficient $\lambda$ }}\hspace{3mm}
In Eqn.\ref{degree}, we use a coefficient $\lambda$  to balance the positive degree and the negative degree. We experiment with different $\lambda$ and show the results in Tab.~\ref{tab:lambda}. We can observe that the test accuracy grows steadily as $\lambda$ increases, which illustrates the effectiveness of the negative degree to reduce the confidence of misclassified sample. However, too large $\lambda$ will restrain the effectiveness of the positive degree and result in accuracy drop. The best $\lambda$ is 0.8 for miniImageNet and 0.6 for tieredImageNet.

%
%To choose a suitable ratio of the negative degree to the positive degree, we search the $\lambda$ from zero and increase it by 0.2 repeatedly. The test accuracies with different $\lambda$  are shown in Fig.~\ref{tab:lambda}. It can be seen that the test accuracy grows steadily as $\lambda$ increases. The best $\lambda$ is 0.8 for miniImageNet and 0.6 for tieredImageNet. When $\lambda$ get very large, it will restrain the effectiveness of the positive degree, thus leading to accuracy drop. So we stop searching after $\lambda=1.0$.

%------------------------------------------------------------------------
\section{Conclusion and Future Work}

In this paper, we propose a novel Progressive Cluster Purification (PCP) method for transductive few-shot learning. The proposed PCP can progressively purify the cluster of each prototype by exploring the semantic interdependency in the individual cluster space. Specifically, our method explicitly exploit the manifold structure of each semantic clusters to refine the prototype, which enhance the ability to represent the real distribution of semantic clusters. And our model is extremely flexible to be repeated several times to progressively purify the clusters. On two challenging benchmarks, the proposed PCP achieves the state-of-the-art results. In future work, we intend to extend the ternary relation among samples to explore manifold structure of a semantic cluster instead of pair-wise relation.

%The proposed PCP can explore the fine-grained semantic interdependency
%in the individual cluster space of each prototype, which leads to more trustworthy predictions for test samples. Specifically, a feature extraction network first extracts discriminative visual features from input images. Then an inter-class classification module partitions the query set into several clusters with prototypes. Moreover, an intra-class transduction module infers relation between samples and refines the prototype to improve its ability to represent the intra-class variability. With these refined prototypes, we can remeasure all the test instances to purify each cluster. We achieve the state-of-the-art results on both miniImageNet and tieredImageNet datasets. We conduct several ablation experiments and demonstrate the effectiveness of each module of the PCP. In future work, we intend to extend the intra-class transduction module by exploiting ternary relation among samples to explore manifold structure of a semantic cluster instead of pair-wise relation.

{\small
\bibliographystyle{ieee}
\bibliography{egbib}
}

\end{document}